# BERT Transformer model for Detecting Arabic GPT2 Auto-Generated Tweets

**Fouzi Harrag, Maria Debbah**
Computer Sciences Department
Ferhat Abbas University, Setif 1
Setif, Algeria
fouzi.harrag,maria.debbahg@univ-setif.dz

**Kareem Darwish, Ahmed Abdelali**
Qatar Computing Research Institute
Hamad Bin Khalifa University (HBKU)
Doha, Qatar
kdarwish,aabdelali@hbku.edu.qa

## Abstract

During the last two decades, we have progressively turned to the Internet and social media to find news, entertain conversations and share opinion. Recently, OpenAI has developed a ma-chine learning system called GPT-2 for Generative Pre-trained Transformer-2, which can pro-duce deepfake texts. It can generate blocks of text based on brief writing prompts that look like they were written by humans, facilitating the spread false or auto-generated text. In line with this progress, and in order to counteract potential dangers, several methods have been pro-posed for detecting text written by these language models. In this paper, we propose a transfer learning based model that will be able to detect if an Arabic sentence is written by humans or automatically generated by bots. Our dataset is based on tweets from a previous work, which we have crawled and extended using the Twitter API. We used GPT2-Small-Arabic to generate fake Arabic Sentences. For evaluation, we compared different recurrent neural network (RNN) word embeddings based baseline models, namely: LSTM, BI-LSTM, GRU and BI-GRU, with a transformer-based model. Our new transfer-learning model has obtained an accuracy up to 98%. To the best of our knowledge, this work is the first study where ARABERT and GPT2 were combined to detect and classify the Arabic auto-generated texts.

## 1 Introduction

In recent years, social media has become an important source of news and information for users in the Arab world, with approximately 63% of Arab youth indicating that they turn to social media first for news (Radcliffe and Bruni, 2019). Due to the lack of editorial checks, fabricated content and misinformation are abundant on different social media platform. For example, recent advances in image and video processing have given rise to so-called "deepfakes". Similarly, large generative language models, such as GPT-2 and GPT-3, have been used to generate increasingly human-like natural text. Such "deepfake" texts can be used by bots, such as Twitter bots, to generate sufficiently diverse content in their timelines to evade detection on social media platforms. We define deepfake texts as "automatically generated string sequences, with or without human seed input, that could fool a human into thinking that was generated by another human.[1]
In this paper, we focus on the problem of detecting such deepfake texts. Specifically, we employ contextual embeddings, which have led to large improvements for a variety of NLP tasks such as part-of-speech tagging, question answering, and text classification (Torfi et al., 2020) without the explicit need for feature engineering. These advances in combination with Deep Learning (DL) enabled the effective detection of auto-generated texts. The trend now is to use deep learning models to solve old machine learning problems including text classification without the need for feature extraction. Deep learning approaches such as Convolutional Neural Networks (CNNs) and Recurrent Neural Networks (RNNs) are now applied to enhance the performance of auto-generated text detection models (Iqbal and Qureshi, 2020). Despite theirs success, they still suffer from the lack of annotated data for training. GPT-2 (Rad-ford et al., 2019), and BERT (Devlin et al., 2019a) are among new approaches recently considered for overcoming this issue. A language understanding model learns contextual and task-independent representations of terms. A huge amount of texts obtained from large corpora is used to build generic (multilingual multi-application) models. Numerous deep learning text detection methods have been pro-posed to help readers determine if a piece of text has been authored by machine or human. The results of employing such language models are very effective in the detection of misinformation and propaganda by providing mechanisms to reveal the nature of news articles and to ensure the authenticity of its source (Dhamani et al., 2019). In this paper, we propose a method for generating deepfake text using GPT2 and to consequently identify such texts using a pre-trained BERT model. As far as we know, it is the first time that such combined approach is proposed for Arabic Language. The outputs of the generative pre-trained language model will be used as inputs for the transfer-learning model that will detect if the input text is coming from human or machine. Our experiment results show that using the pre-trained BERT model outperforms RNN models in detecting Arabic deepfake texts.

---

[1] https://www.wired.com/story/ai-generated-text-is-the-scariest-deepfake-of-all/

The contributions of this work are:

- Generating a new Arabic dataset using GPT2-Small-Arabic.
- Proposing an efficient model to detect and identify auto-generated text using AraBERT.
- Comparing with different RNN-based baseline models.

## 2 Related Work

Several studies have been carried out on the classification and detection of social bots. In this section, we will provide an overview of relevant related works. This can provide a basic understanding on the state-of-the-art of social bot detection methods and approaches. Almerekhi and Elsayed (2015) proposed a system for the detection of automatically generated Arabic tweets. They evaluated their system using 3.5k labelled tweets randomly sampled and labeled by Crowd Flower. A set of features including formality, structural, tweet-specific, and temporal features has been extracted from Arabic tweets to train three classification algorithms, namely: Support Vector Machines (SVM), Naive Bayes, and Decision Trees. Their results show that classification based on individual categories of features outperform the baseline unigram-based classifier in terms of classification accuracy.

Varol et al. (2017) proposed a framework for bot detection on Twitter. They presented a machine learning based system used for the extraction of a collection of features from six different classes. The features were used in training robust models to identify bots. They evaluated the performance of their detection system using two different datasets: manually-annotated Twitter accounts dataset and a pub-lic honeypot dataset. Their analysis on the contributions of different feature classes suggests that user metadata and content features are the two most valuable sources of data to detect simple bots.

Darwish et al. (2017) looked at the problem of detecting propaganda accounts on Twitter. They crafted three different kinds of features, namely interaction features (ex. retweets and mentions), lexical diversity, and stylistic features (ex. offensiveness and sentiment). Using an SVM classifier, they found that interaction features where the most effective features for detecting propaganda accounts, and stylistic features benefited overall classification.

Akyon and Kalfaoglu (2019) presented a system for the detection of fake and automated accounts. Their system is tackling the problem of using a binary classification model. They constructed a dataset of 1,400 accounts from different countries. There were an equal number of real and automated accounts in their dataset. They proposed a set of derived features and the most effective features from this set were selected to develop a cost-sensitive genetic feature selection algorithm. Their results showed that neural network methods and SVM achieved the best F1 scores for the detection of automated accounts.

Santia et al. (2019) presented a simple model based on capturing regions to detect bots on Facebook. Features were leveraged in experimentation as bot indicators to separate bots away from the central range of the full feature distributions. They noted that some features that appear to cluster bots tightly are collaborating with high priority in the identification of a user as a human. Their system has been developed and evaluated on public pages as bot detection software using standard evaluation metrics such as F1 score. They found that a small percentage of interacting user population were showing signs to be social bots.

In the work of Kudugunta and Ferrara (2018) work, a deep neural network based on contextual long short-term memory (LSTM) was proposed. The architecture of the system exploits content and metadata features to detect bots at the tweet level. They used the features extracted from user's profile metadata and tweet text as inputs to the LSTM. They also proposed a very useful data augmentation technique for generating large labeled datasets based on synthetic minority oversampling. They showed that their model could identify bots from humans with a high accuracy. They also achieved a perfect classification accuracy by applying the same architecture to account-level bot detection.

Wei and Nguyen (2019) developed an RNN (biLSTM) word embeddings based model to discriminate Twitter bots from human accounts without requiring any prior knowledge or handcrafted features. By comparison with similar bot detection systems, their experiments on the Cresci-2017 dataset showed that their proposed system achieved reasonable performances. They also confirmed the ability of using biLSTM with word embeddings in their model to detect phishing email, webpages or SMSs.

In the work of Cai et al. (2017), a behavior enhanced deep model for bot detection was proposed and applied to detect bots under two deep learning frameworks. By fusing content and behavior information, the aim of the proposed model is to capture the latent features. Using a honeypot method, they collected a public dataset for theirs experiments using a CNN network. Their model was investigated by varying the number of filters and the filter width and number of hidden units in the hidden layer. Their proposed system achieved the highest F1 score compared to other state of the art systems.

## 3  Data Preparation

### 3.1  Collecting the Dataset

In this section, we focus on creating a dataset for detecting deepfake text. To the best of our knowledge, no such dataset exists. The closest dataset to our needs is that of Almerekhi and Elsayed (2015), which includes 3,503 tweets that are manually labeled as "human content" (1,559 tweets) or "bot content" (1,944 tweets). Upon inspecting their dataset, much of the bot content includes verses from the Quran, the Muslim holy book, proverbs, and sexual content. Though such are being distributed by bots, they don't fit the definition of deepfake text, which entails human-like texts that were automatically generated by machines with or without seed input. Thus, we needed to create a new dataset.

To prepare our dataset, we utilized the 1,559 tweets in the dataset of Almerekhi and Elsayed (2015) that were labeled as "human content", as we are sure are authored by humans. Due to the limited number of tweets in this set, we expanded the set by crawling the timeline tweets of the users in the set. This is based on the simplifying assumption that accounts that produced human content are unlikely to produce bot content or deepfake text. It is noteworthy that some of the accounts from which the tweets originated were in fact deleted or suspended or individual tweets themselves were deleted. Thus, we removed deleted tweets and we were not able to get the timelines of deleted or suspended accounts. After expansion, the number of tweets in our dataset increased to 4,196. We henceforth refer to this expanded set as the human generated set. We normalized the text of the tweets by removing URLs, splitting hashtags, replacing user mentions by USER, removing non-Arabic characters and punctuation marks, and removing diacritics, which are short vowels that are optionally written in Arabic text.

Given our human generated set, we needed a contrasting deepfake text set. To do so, we used the GPT2-Small-Arabic[2] model to auto-generate sentences based on our human generated set. Generally, GPT-2 is a pre-trained model that generates synthetic sentences by predicting the next word based on previous words (Radford et al., 2019). Similarly, GPT2-Small-Arabic was pre-trained on an Arabic Wikipedia dump (around 900MB of text) using the Fastai2 library (Abed, 2020). Unlike other transfer learning-based models that need 2 stages for training (pre-training and fine-tuning stages), GPT-2 like models do not require fine-tuning (Ma, 2019). We used GPT2-Small-Arabic to generate deepfake text that is seeded using the tweets in the human generated set. We generated a set of deepfake texts using GPT2-Small-Arabic, where the generated texts ranged in length between 15 and 35 words. The set is composed of 3,512 deepfake texts. We henceforth refer to this set as the deepfake set. Table 3 shows an example of a tweet from the human generated set and the corresponding deepfake text that was generated from it.

---

[2] https://huggingface.co/akhooli/gpt2-small-arabic

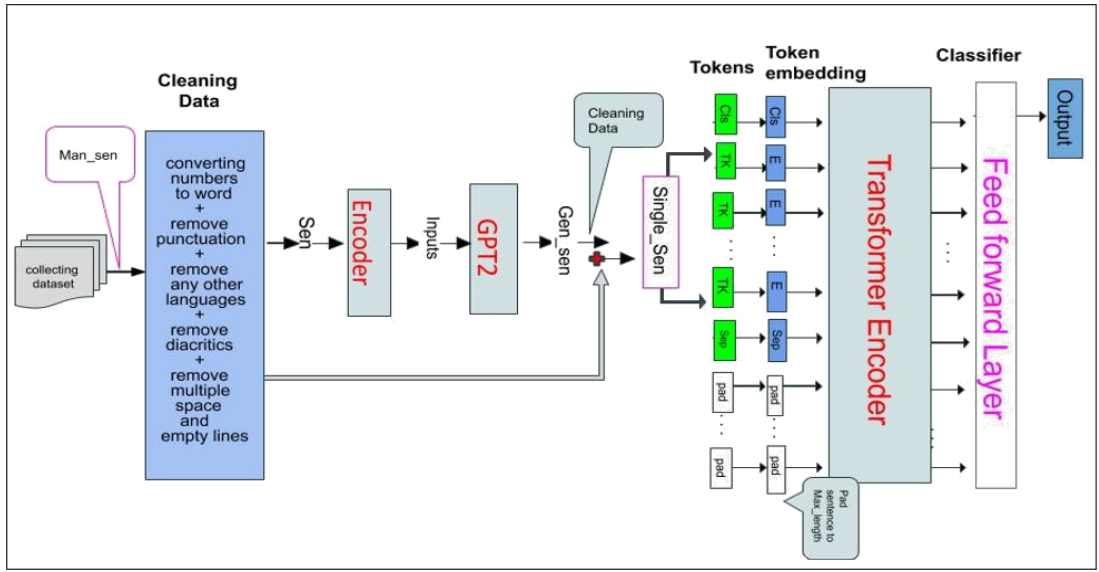

**Figure 1**: Overall architecture.

| Abbreviation | Signification |
| --- | --- |
| Sen | Sentence. |
| Man_Sen | Manual sentences. |
| Gen_Sen | Generated sentences. |
| Pad | Padding. |
| Tk | Tokens. |
| E | Embedding token |
| Max_length | Maximum length of all the sentences. |
| CIS | For classification tasks, we must append the special [CLS] token to the beginning of every sentence. |
| Sep | We need to append the special [SEP] token at the end of every sentence. |

**Table 1**: Architectural elements.

## 4 Experimental Setup

We combined the human generated and deepfake sets, and we randomly divided the combined dataset into 80/20 training and test splits. We utilized 4 competitive deep learning Recurrent Neural Network (RNN) baselines to compare to our proposed transformer-based model. Figure 1 showcases the entire architecture of our proposed solution with the explanation of the elements therein in Table 1.

| parameter | LSTM | GRU | biLSTM | biGRU |
|---|---|---|---|---|
| Input layer size | 50 | | | |
| RNN layer size | 128 | | | |
| Dense layer size | 400 | | | |
| Learning rate | 0.001 | | | |
| Batch size | 100 | | | |
| Dropout (dense layer) | 0.4 | 0.5 | 0.3 | 0.5 |
| Epochs | early stopping with patience of 4 | | | |

**Table 2:** RNN model parameters.

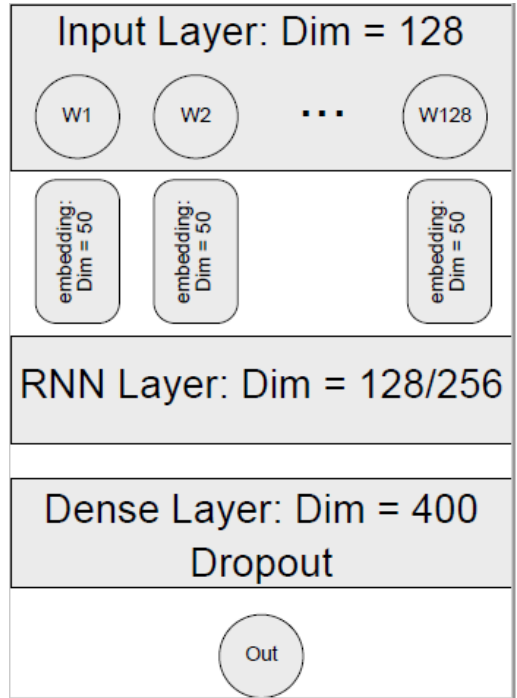

**Figure 2:** RNN baseline architecture.

### 4.1 Baseline

For the baselines, we used 4 different RNN models. As shown in Figure 2, all the models have a cascade of an input layer of size 128, randomly initialized word embeddings of size 50, an RNN layer of size 128 (or 256 for bidirectional), a dense layer of size 400, and an output layer with softmax activation. We used 4 different RNN layers, namely Long Short Term Memory (LSTM), Gated Recurrent Unit (GRU), bidirectional LSTM (biLSTM), and bidirectional GRU (biGRU). LSTMs have the advantage of effectively dealing with the vanishing gradients problem. Bidirectional GRUs and LSTMs make better use of the training data as the data is traversed twice, typically leading to improved performance (Siami-Namini et al., 2019). Table 2 lists the model parameters.

### 4.2 Our BERT-based Method

Recently, deep contextualized language models such as BERT (Bidirectional Encoder Representations from Transformers) (Devlin et al., 2019b) resulted in major improvements for many NLP classification and language understanding tasks. For our proposed method, we fine-tuned AraBERT (v 0.1) for the classification task (Antoun et al., 2020). AraBERT is pre-trained on an identical architecture to BERT, namely an encoder with 12 Transformer blocks, hidden size of 768, and 12 self-attention heads. It is trained on a large Arabic news corpus containing 8.5M articles composed of roughly 2.5B tokens, and it uses SentencePiece (BP) word segmentation. AraBERT has been shown to yield better results than multilingual BERT from Google, which is trained on Arabic Wikipedia only (Antoun et al., 2020). We used the PyTorch[3] implementation by HuggingFace[4] as it provides pre-trained weights and vocabularies, and we carried out 4 fine-tuning epochs.

| | |
|---|---|
| Human generated | اليوم بعد الفجر رأيت رؤيا بمحمد إن شاء الله إنها خير |
| Deepfake | اليوم بعد الفجر. تم إطلاق على القوات الاسرائيلية على الفور، وتم إطلاق النار على طائرة إسرائيلية ثنائية الجنسية في 26 مارس، مما أسفر عن مقتل أربعة وإصابة خمسة أشخاص. وقال |

**Table 3:** An example of GPT2-Small-Arabic generated text

| | Accuracy | Precision | Recall | F1-Score |
|---|---|---|---|---|
| LSTM | 95.0 | 92.8 | 96.8 | 94.8 |
| Bi-LSTM | 96.3 | 96.9 | 95.2 | 96.0 |
| GRU | 94.7 | 91.2 | 98.2 | 94.6 |
| Bi-GRU | 95.9 | 93.7 | 97.8 | 95.7 |

**Table 4:** Results for RNN based baseline models

## 5 Experimental Results

### 5.1 Baseline Results

Table 4 shows the results of dataset for our four RNN Baseline models, namely LSTM, biLSTM, GRU and biGRU. BiLSTM and biGRU were the best performers. BiLSTM has the highest balanced perfor-mance, compared to others models. Further, using bidirectional RNN, either biLSTM or biGRU, instead of unidirectional RNNs had a noticeable positive improvement on the results.

### 5.1.1 AraBERT Results

Table 5 shows the AraBERT results. The results clearly indicate the ability of transfer learning models to detect auto-generated texts with a high degree of accuracy. In addition, it is also clear through these experiments to what extent the hyperparameters impact the performance of these classifiers. Also, the AraBERT model significantly outperformed all baseline models. The improvements compared to the baseline results are +3.7% for LSTM, 4.0% for GRU, +2.4% for biLSTM, and +2.8% for biGRU. To better visualize the improvements, Figure 3 compares the results.

## 6 Conclusion

In this paper, we aimed to automatically detect deepfake text. We created a dataset of human generated tweets and deepfake tweets that we generated using GPT2-Small-Arabic. For classification, we experimented with different RNN models and compared them to AraBERT, a pre-trained BERT model for Arabic. The results show that we can effectively distinguish between human generated and machine generated tweets with high accuracy (98.7%) using a fine-tuned AraBERT model, which outperforms RNN models. For future work, we plan to use explainability tools to determine the reasons for the success of the AraBERT model. We suspect that the model is able to detect deepfakes due to repetitions in the text, which are a common artifact of automatically generated text, or the style of language of GPT2 generated text.

---

[3] https://pytorch.org/

[4] https://github.com/huggingface/transformers

| Accuracy | Precision | Recall | F1-Score |
|---|---|---|---|
| 98.7 | 98.9 | 98.5 | 98.7 |

**Table 5:** Results for AraBERT model

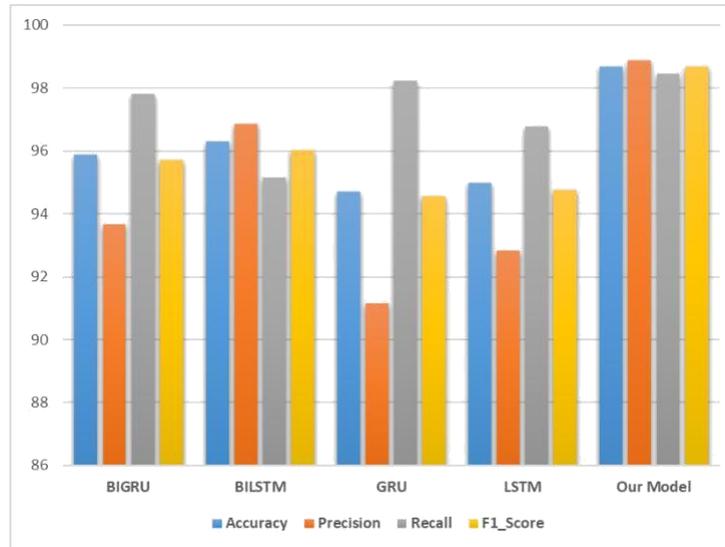

**Figure 3:** Improvement of our model compared to RNN baseline models